\documentclass[a4paper,fleqn]{cas-sc}

\renewcommand{\printorcid}{}

\usepackage[T1]{fontenc}
\usepackage[utf8]{inputenc} 
\usepackage{amsmath}
\usepackage{amsthm} 
\usepackage{booktabs} 
\usepackage{multirow} 
\usepackage{amssymb} 
\usepackage{float}
\usepackage{booktabs}
\usepackage{tabularx}
\usepackage{longtable}
\usepackage{xcolor}
\usepackage{caption}
\usepackage{adjustbox}
\usepackage{float}
\usepackage[numbers,sort&compress]{natbib}

\restylefloat{table}
\restylefloat{figure}
\usepackage{graphicx,subfig}
\DeclareGraphicsExtensions{.pdf,.jpeg,.png,.jpg,.svg, .tiff}
\graphicspath{{myfigs/}}

\newtheorem{def1}{Definition}

\def\tsc#1{\csdef{#1}{\textsc{\lowercase{#1}}\xspace}}
\tsc{ICT}
\tsc{IoT}

\begin{document}
\let\WriteBookmarks\relax
\def\floatpagepagefraction{1}
\def\textpagefraction{.001}
\shorttitle{Systemic formalisation of Cyber-Physical-Social System(CPSS)}
\shortauthors{BA Yilma et~al.}

\title [mode = title]{Systemic formalisation of Cyber-Physical-Social System (CPSS): A Systematic Literature Review
}

\author[1,2]{Bereket Abera Yilma}
\ead{bereket.yilma@list.lu}
\address[1]{Luxembourg Institute of Science and Technology (LIST), 5, Avenue des Hauts-Fourneaux, L-4362, Esch-sur-Alzette, Luxembourg}
\address[2]{Université de Lorraine, CNRS, CRAN, F-54000 Nancy, France}
\author[2]{Hervé Panetto}
\ead{herve.panetto@univ-lorraine.fr}
\author[1]{Yannick Naudet}
\ead{yannick.naudet@list.lu}

\begin{abstract}
The notion of Cyber-Physical-Social System (CPSS) is an emerging concept developed as a result of the need to understand the impact of Cyber-Physical Systems (CPS) on humans and vice versa. This paradigm shift from CPS to CPSS was mainly attributed to the increasing use of sensor enabled smart devices and the tight link with the users. The concept of CPSS has been around for over a decade and it has gained an increasing attention over the past few years. The evolution to incorporate human aspects in the CPS research has unlocked a number of research challenges. Particularly human dynamics brings additional complexity that is yet to be explored.  The exploration to conceptualise the notion of CPSS has been partially addressed in few scientific literatures.  Although its conceptualisation has always been use-case dependent. Thus, there is a lack of generic view as most works focus on specific domains. Furthermore the systemic core and design principles linking it with the theory of systems are loose. This work aims at addressing these issues by first exploring and analysing scientific literatures to understand the complete spectrum of CPSS through a Systematic Literature Review (SLR). Thereby identifying the state-of-the-art perspectives on CPSS regarding definitions, underlining principles and application areas. Subsequently, based on the findings of the SLR, we propose a domain-independent definition and a meta-model for CPSS, grounded in the Theory of Systems. Finally a discussion on feasible future research directions is presented based on the systemic notion and the proposed meta-models.

\end{abstract}


\maketitle
\section{Introduction}
\label{sec:introduction}
Cyber-Physical-Social System(CPSS) is an emerging research topic resulting from the introduction of a Social dimension to the existing Cyber-Physical System (CPS) research. The notion of CPS was originally derived from an engineering perspective with the support of the US National Science Foundation (NSF)~\cite{koubaa2009vision, gunes2014survey}. The objective of CPS was mainly controlling and monitoring physical environments and phenomena via the integration of sensing, computing, and actuating devices. In parallel the notion of Internet of Things (IoT) which originates in the late 90's\footnote{The earliest quote about Internet of Things in  a presentation  from  Kevin  Ashton, 1999,  MIT  Auto-ID  Center,  reported  in  Forbes  in  2002 (see https://www.forbes.com/global/2002/0318/092.html5f7b1b353c3e)} took off as a paradigm with the support of the European Commission from a Computer Science perspective~\cite{gubbi2013internet}.  The emergence of IoT played an indispensable role for the orchestration of the physical and cyber systems with the goal of connecting tools and electronic equipment to the Internet and develop a network of computers and objects.

Despite their initial philosophical difference, IoT and CPS share many similarities hence they have been used sometimes interchangeably without a clearly defined demarcation \cite{jeschke2013everything}. There is however a fundamental difference that should be highlighted: the fact that a CPS refers to a particular type of systems explicitly, while IoT refers at the same time to the concept, the system formed by all the connected devices and a particular system of interconnected objects. As a system, a CPS typically collects and controls information about phenomena from the physical world through networks of interconnected devices, in order to achieve its objective \cite{lee2008cyber}. In the development of both paradigms, humans were originally assumed as external entities interacting with these systems. 

Over the years the increasing popularity of smart devices and their significant role in the daily life of their users has led CPS systems to consider humans as a multifaceted source of information (i.e. human sensors)\cite{7865933,su2017incentive}. Subsequent research studies have started to incorporate humans in CPS research. This trend in particular has uncovered the importance of humans' centrality for the development of CPS which was then recognised by the Human-in-the-Loop (HitL) CPS paradigm~\cite{nunes2015survey,Sowe2016a,munir2013cyber}, where humans are intrinsic actors of the system.  Different techniques have then been used to introduce human actors in CPS, paving the way for the foundation of Cyber-Physical-Social System(CPSS) \cite{Yilma,Yilma2019}. Over the past decade different researchers used different terminologies to refer to the integration of the human aspect in to CPS projecting different conceptualizations . For instance in \cite{Sowe2016,Smirnov2017a,kumar2017securing,quintas2016information,6685535,wu2014treatment,7961580,Zhang2017,tsoukalas2017active,haggi2019review}  Cyber-physical-human systems (CPHS) was used, being defined as \textit{``a system of interconnected systems (computers, cyber-physical devices, and people) “talking” to each other across space and time, and allowing other systems, devices, and data streams to connect and disconnect.''} Alternatively Human-cyber-physical system(HCPS) was also used in some works perceiving human elements as physical entities interconnected with CPS, being controlled and coordinated (rather passively) by cyber
systems in the whole system \cite{liu2020human}  In \cite{7428352} the concept of Cyber-Physical-Social-Thinking hyperspace (CPST) was introduced for geological information service system. These works define CPSS as \textit{``a system deployed with emphasis on humans, knowledge, society, and culture, in addition to Cyber space and  Physical space. Hence, it can connect nature, cyber-space, and society with certain rules.''} where as CPST is established through the mergence of a new dimension of thinking space into the CPS space. The thinking space is \textit{a high-level thought or idea raised during the intellectual activities of people.} These works visualize the \textit{intellect} of humans separately from the Social aspect of CPSS  as Thinking space. On the other hand the term Social-Cyber-Physical-Systems (SCPS) was also used in \cite{Kannisto2016,Xu2017,inproceedings, horvath2012beyond}. The commonly shared understanding of SCPS was \textit{``complex socio-technical systems, in which human and technical aspects (CPS) are massively intertwined.''} as defined by \cite{horvath2012beyond}. According to this definition the awareness of SCPS extends to the intangibles of social
context, which includes social culture and norms, personal beliefs and attitudes, and informal institutions of social interactions. Finally, the term Cyber-Physical Human-Machine system was also used in \cite{chen2017guest}.

Nowadays the acronym CPSS is being widely used in various application areas. Smart Cities, Smart Homes, Schools, Offices, Museums, and medium to large scale Industries are among the main sectors, where applications of the CPSS notion has gained momentum \cite{Yilma,Yilma2019,Huang2020,Zhong2020,Wang2020a,Shen2020,Peng2020,Xu2020,Peng2020a, Zhang2020,Zhu2020,Liu2020,Dai2020,Wang2019a,li2019multi,FANTINI2019133, Wang201942,8409112,CASSANDRAS2016156, de2017cyber,amin2020hotspots}. 
Despite the advances made to integrate human aspects in CPS, the development of CPSS research is still in its infancy. 
There are still no reliable approaches to harmoniously model the social part together with cyber and physical. Additionally the hardly predictable and complex nature of humans by itself brings additional complexity to the system. Humans are social creatures and the term "Social" carries a broader meaning as it reflects emotional, cognitive and behavioral aspects of a human \cite{Yilma2019} which are deemed as the three layers of human interaction responses  \cite{Norman02,norman2014, PERUZZINI2018105600}. Moving forward towards a harmonious integration of social aspects, the CPSS research should take a holistic approach on the synergy between social aspects and CPS \cite{Yilma2019}. However, today, the commonly adopted conceptualisations of CPSS often capture partial characteristics of the social dimension. 

In this sense, there is a need for reaching a common understanding on the concepts of CPSS to guide in the development of technologies. In recent years some works carried out literature reviews with the objective to provide a view on CPSS development, challenges and application areas \cite{ gunes2014survey,nunes2015survey,de2017cyber,zhou2019cyber, 7434330,haggi2019review,BAVARESCO2019109365,Wang201942, zhang2018cyber}. While some works \cite{gunes2014survey,nunes2015survey} offer a brief overview on the conceptualisation of CPSS, almost all the rest tend to focus on summarising the application areas and the particular problems addressed in the literature. Although a seemingly common understanding may have been reached in some domains and, in some cases, more general concepts were introduced \cite{Yilma2019}, there is a lack of a uniform systemic ground and domain-independent view of CPSS. In general to the best of our knowledge, there is a lack of consensus regarding the conceptualisation of CPSS and its characteristics considering commonalities how they are treated in different domains in the scientific literature. Therefore, this work aims at exploring the state-of-the-art regarding Cyber-Physical-Social System (CPSS). We put special focus on identifying definitions and characteristics of CPSS and  how the social and human aspects are integrated in current research trends.

To this end, we conduct a Systematic Literature Review (SLR) according to the guidelines proposed by Kitchenham \cite{kitchenham2004procedures}. 
The contributions of this work can be summarised as: (1) to provide a view on the state-of-the-art perspectives of CPSS considering how they are defined, and how social aspects are depicted within various domains in CPSS literature. 
(2) to establish a common ground towards a unified definition of CPSS through a systemic formalisation and a meta-model proposal, and (3) propose a future research direction to support the integration of social aspects in CPSS research.

This paper is organised as follows. Section \ref{sec:research_context} introduces the research context. Section \ref{sec:research_context:sota} describes the procedure followed to perform the SLR and the results obtained from the descriptive analysis. Section \ref{sec:results} presents the state-of-the-art perspectives on CPSS, by discussing definitions and how the social aspect is conceptualised in literature. Section \ref{sec:CPSS_formalisation} presents the Systemic formalisation of CPSS by establishing a link with the theory of systems. Section \ref{sec:CPSS_Future} covers a discussion on feasible future research direction. Finally, Section \ref{sec:conclusion} presents a concluding discussion.

\section{Research Context}
\label{sec:research_context}

To understand what a CPSS is, how it has evolved, reflect on state-of-the-art perspectives and propose future research directions, this study formed by SLR tries to answer the following key research questions:  

\begin{itemize}
    \item How is a CPSS defined?
    \item How is the Social dimension (i.e. human aspect) conceptualised in current CPSS research?
    \item What are the application areas of CPSS?
    \item What are the main issues and challenges in the current CPSS research? (mainly due to the active involvement of humans).
    \item What can be made to address these challenges?
\end{itemize}

\section{Systematic Literature review}
\label{sec:research_context:sota}

\subsection{Paper Selection Procedure}
In this subsection we present the details of our paper selection procedures (i.e. Database selection, Keyword and Search strings definition and paper filtering steps). The initial screening procedure was carried out by a single author and later verified by two independent authors. 
\subsubsection{Database Selection}
\label{sec:method:literature_review:database}
In order to cover all relevant studies that could potentially answer the above mentioned key research questions we searched for papers by querying the following digital libraries:  ACM, Scopus, IEEE\_Xplore, Taylor \& Francis Online, Wiley and Springer. We selected six databases taking into account that a minimum of four is deemed sufficient to perform a robust literature search \citep{2010_Kitchenham}. It is worth mentioning that among the six queried databases Scopus is known to be an extensive abstract and citation database that gathers papers from several peer-reviewed journals. The papers retrieved from this database come from diverse publishers such as Elsevier, Springer, Taylor \& Francis Online, and IEEE. It is expected that this will provide more robustness to the search. Normally, duplicated papers are expected to appear from this search and they are removed.
Each database has its own syntax to write queries. Hence, the search strings described in Section \ref{sec:method:literature_review:keywords} are slightly modified for each database to obtain the expected output.

\subsubsection{Keywords and search strings definition}
\label{sec:method:literature_review:keywords}
The keywords used in this study were defined based on an iterative process, which is described as follows. First we queried the digital libraries with the search string  \textit{Cyber Physical Social System} denoted by S$_{1}$. A total of 431 papers were retrieved.  From these papers, we extract the most used keywords (i.e. repeated more than five times) and the most repeated terms (i.e. repeated more than fifteen times) in their titles and abstract. This was done by downloading paper's metadata (i.e. title, year of publication, authors, abstract and keywords). Next, we perform a data mining on the extracted metadata in order to identify the relevant keywords using VOSviewer software \cite{van2010software}. VOSviewer is used to construct and visualise co-occurrence networks of important terms extracted from the metadata. Additionally  a manual analysis of the metadata was carried out in order to identify relevant keywords. Combining the two we identified twelve additional keywords and redefined the search string  S$_{1}$ to S$_{2}$.
\begin{itemize}
    \item S$_{1}$: \textit{\{Cyber Physical Social System}\}
    \item S$_{2}$: \textit{\{Cyber Physical Social system}\}, \textit{\{Human Cyber Physical System}\}, \textit{\{Socio Cyber Physical System}\}, \textit{\{Social IoT}\}, \textit{\{Cyber physical Human System}\}, \textit{\{Social Cyber Physical System}\}, \textit{\{Human in the Loop CPS}\}, \textit{\{Cyber Physical Social Thinking}\}, \textit{\{Cognitive IOT}\}, \textit{\{Human Centered  IoT}\}, \textit{\{Human Centered  CPS}\}, \textit{\{Human in the Mesh}\}.
\end{itemize}
Querying the databases with S$_{2}$ we retrieved a total of 705 papers. Table  \ref{tab:search_results} summarises the total number of papers obtained from each database per search string. 

\begin{table}[!h]
\caption{Number of papers retrieved from each database, per search string.}
\centering
\begin{tabular}{@{}lcccc@{}}
\toprule
 \multicolumn{1}{l}{\textbf{Publisher Databases}}
 & \multicolumn{1}{l}{\textbf{Search string S$_{1}$}} &                   \multicolumn{1}{l}{\textbf{Search string S$_{2}$}} \\ \midrule
\textbf{ACM} & 9 & 114  \\
\textbf{Scopus} & 248 & 260  \\
\textbf{IEEE\_Xplore} & 67 & 189  \\
\textbf{Taylor \& Francis} & 4 & 4  \\
\textbf{SpringerLink} & 99 & 101  \\
\textbf{Wiley} & 4 & 37  \\ \midrule
\textbf{Total} & \textbf{431} & \textbf{705}  \\ \bottomrule
\end{tabular}
\label{tab:search_results}
\end{table}

\subsubsection{Papers Filtering}
\label{sec:method:literature_review:filtering}
After retrieving 705 papers from the the database search,
we included 34 additional papers using the \textit{“snowball sampling”} technique \cite{wohlin2014guidelines}. In the \textit{“snowball sampling”} we considered the referrals of CPSS approaches made by experts, as well as the most cited papers in the existing surveys and reviews. We did not restrict our search with publication year  and citation impact in order not to miss out those papers with relevant definitions but not cited  enough as they were published recently  or due to the narrow domain of interest of the papers. Therefore we considered  all papers offering definitions of CPSS regardless of publication time and citation impact.   A total of 739 publications were identified at the end of this sampling phase. Subsequently the selection of papers to be analysed and included in the study was done in a two step filtering mechanism.  In the first filtering step represented by $F_{1}$ the metadata of each paper was screened through a set of inclusion and exclusion criteria. The second filtering step involves reading the full text of the remaining papers. To select which papers are to be considered in this study an additional set of inclusion and exclusion criteria  were applied represented by $F_{2}$. The two step filtering mechanism is described in table \ref{tab: F1_inclusion_exclusion_criteria},

\begin{table}[!h]
\caption{Inclusion and exclusion criteria used to select the papers. The criteria are organised in two filters, $F_{1}$ and $F_{2}$, the first is related to metadata analysis whilst the second is based on analysing the full text of the papers.}
\centering
\begin{tabularx}{\textwidth}{p{.2\textwidth}p{.4\textwidth}X}
\toprule
\textbf{Filter} & \textbf{Inclusion Criteria} & \textbf{Exclusion Criteria} \\ \midrule
\multirow{3}{*}{$F_{1}$} & Papers written in English & Papers not written in English \\ 
 & Paper with full text access & Paper without full text access \\ 
 & Primary studies  & Literature reviews \\  \midrule
  \multirow{2}{*}{$F_{2}$} & Papers establishing a link between CPS/IoT and social dimensions/ human aspects &  Papers without link between CPS/IoT and social dimensions/ human aspects\\
& Papers containing definitions & Papers that are only about CPS/IoT \\
 & Papers with implicit definitions & \\
 & Papers discussing open challenges in CPSS  &  \\
 & Papers proposing solutions for social dynamics in CPSS  &  \\
  \bottomrule
\end{tabularx}
\label{tab: F1_inclusion_exclusion_criteria}
\end{table}
 

\subsection{Descriptive Analysis of Papers}
\label{sec:method:papers_analysis}
The initial search revealed 705 references from the digital libraries, and 34 papers based on the snowball sampling. We then applied filtering on the total of 739 papers guided by a series of inclusion and exclusion criteria as described in table \ref{tab: F1_inclusion_exclusion_criteria}. First we excluded those papers that are not accessible , not written in English and papers that are reviews and surveys. Consequently the number of considered papers dropped to 589. Moving forward, we analysed the rest of the papers, considering their title, abstract and keywords. Thus, the number of considered papers dropped to 427. Moreover, after reading and analysing the full text of the remaining papers according to inclusion and exclusion criteria, we selected 122 of them.  Once the papers are selected, we classify them by the type of publication (e.g. journal article, conference proceedings, etc.), year of publication and country. Table \ref{tab:selection_process} shows the details of the selection process.

\begin{table}[!h]
\caption{The details of paper selection process.}
\centering
\begin{tabular}{@{}lcccc@{}}
\toprule
 \multicolumn{1}{l}{\textbf{Phase}}
 & \multicolumn{1}{l}{\textbf{Total N° of Papers}} \\ \midrule
Total number of paper from digital libraries & 705   \\
N° of papers after snowballing sampling & 739   \\
N° of papers after exclusion based on the paper access, language and type of research & 589   \\
N° of papers after exclusion based on title, abstract and keywords & 427  \\\midrule
\textbf{N° of papers after exclusion based on full text =  N° of included papers}   &\textbf{ 122}   \\ \bottomrule
\end{tabular}
\label{tab:selection_process}
\end{table}

\begin{figure}[!h]
  \centering
  \captionsetup{justification=centering}
  \includegraphics[width=0.4\textwidth]{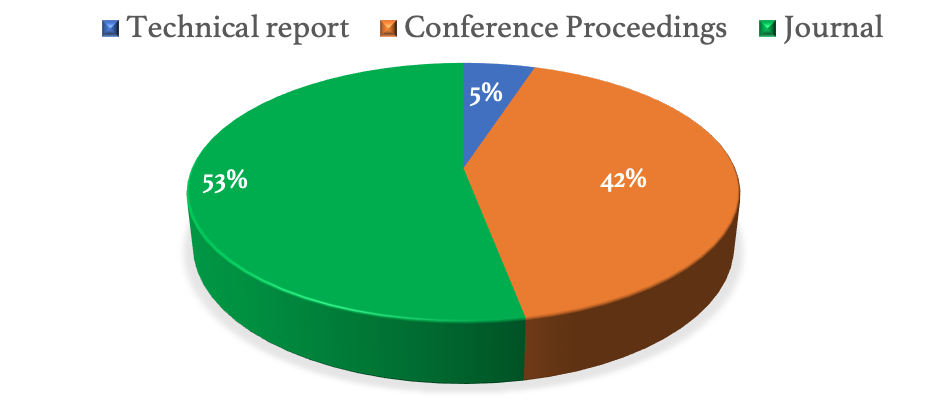}
  \caption{ The type of publications selected in this study.}
  \label{fig:Paper_type}
\end{figure}

The analysis shows that the selected publications are composed of journal articles amounting to 53$\%$ and conference proceedings amounting to 42$\%$. The remaining 5$\%$ were identified as technical reports. Figure \ref{fig:Paper_type} summarises they type of publications selected in this study.

\begin{figure}[!h]
  \centering
  \captionsetup{justification=centering}
  \includegraphics[width=0.6\textwidth]{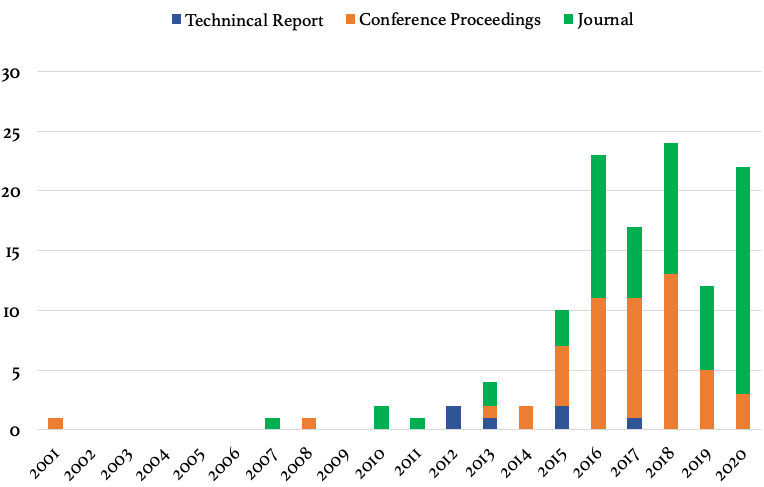}
  \caption{ Number of papers published per year.}
  \label{fig:Paper_years}
\end{figure}

The overall sample considered in this study  constitutes papers published up to May 2020. The
time distribution of the papers published is shown in Figure \ref{fig:Paper_years}. A small fluctuation can be seen between the years 2007 and 2014 in the number of papers. In 2015 a steady growth appeared followed by a sharp increase in 2016 with gradual changes in 2017 and 2018. In the year 2020, the rate of publication only within  the first few moths has almost doubled the previous year. This evolution rationalises the the increasing attention the CPSS research gained over the years.

\begin{figure}[!h]
  \centering
  \captionsetup{justification=centering}
  \includegraphics[width=0.7\textwidth]{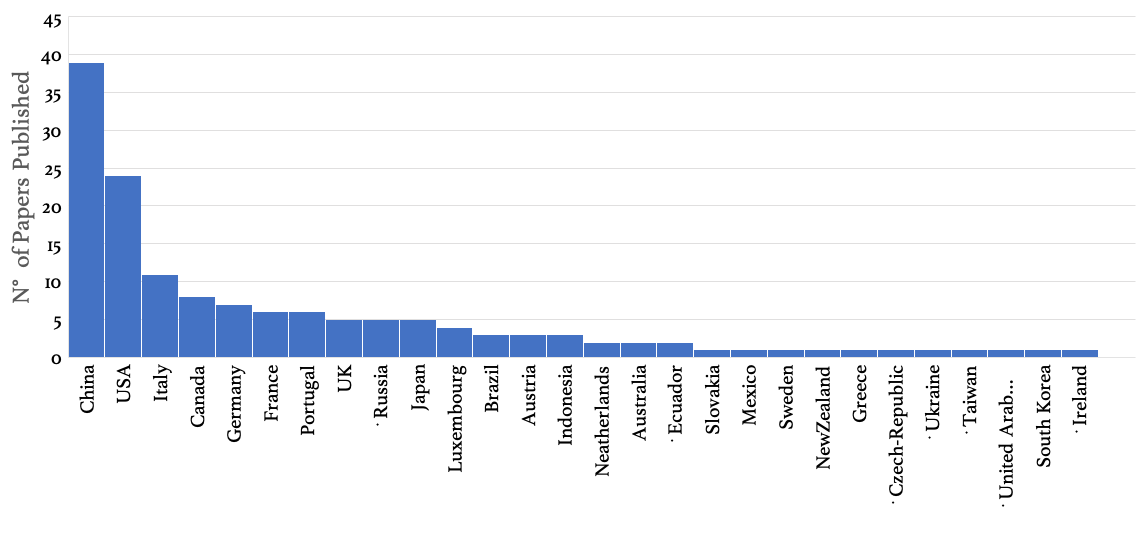}
  \caption{ Number of papers published per Country.}
  \label{fig:Paper_Countries}
\end{figure}

\begin{figure}[!h]
  \centering
  \captionsetup{justification=centering}
  \includegraphics[width=0.6\textwidth]{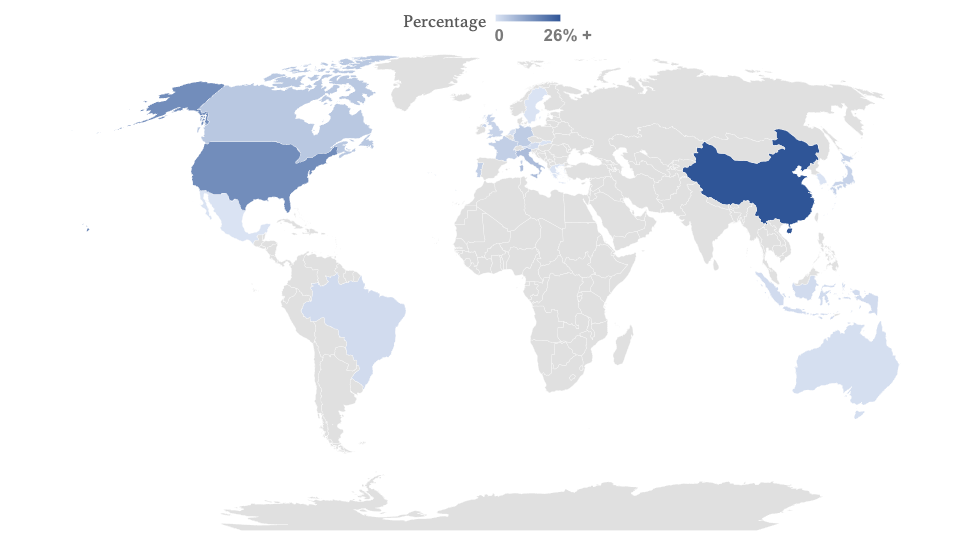}
  \caption{ The development of CPSS research worldwide according to the number of publication.}
  \label{fig:CPSS_worlwide}
\end{figure}

\noindent
Our analysis also revealed the development of CPSS research across nations worldwide. As it can be seen on Figure \ref{fig:Paper_Countries},  \textit{China, USA} and \textit{Italy} are the leading contributors followed by \textit{Canada, Germany and France} in terms of the number of publications produced until May 2020. Figure \ref{fig:CPSS_worlwide} illustrates the development of CPSS research worldwide according to the number of publication produced by countries.
\begin{figure}[!h]
  \centering
  \captionsetup{justification=centering}
  \includegraphics[width=0.5\textwidth]{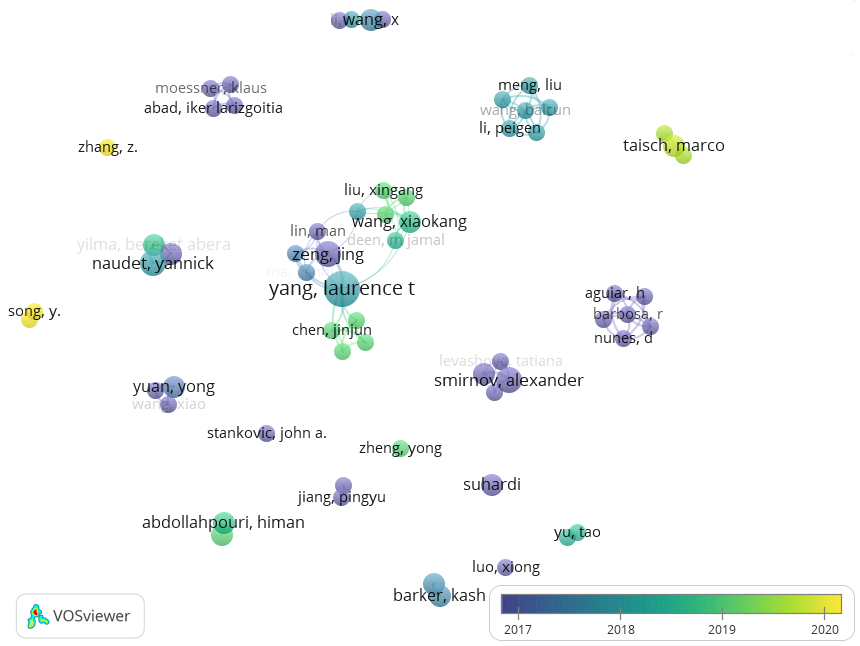}
  \caption{ Co-authorship network of the top 20 authors.}
  \label{fig:Co-authorship_network}
\end{figure}
\begin{figure}[!h]
  \centering
  \captionsetup{justification=centering}
  \includegraphics[width=0.4\textwidth]{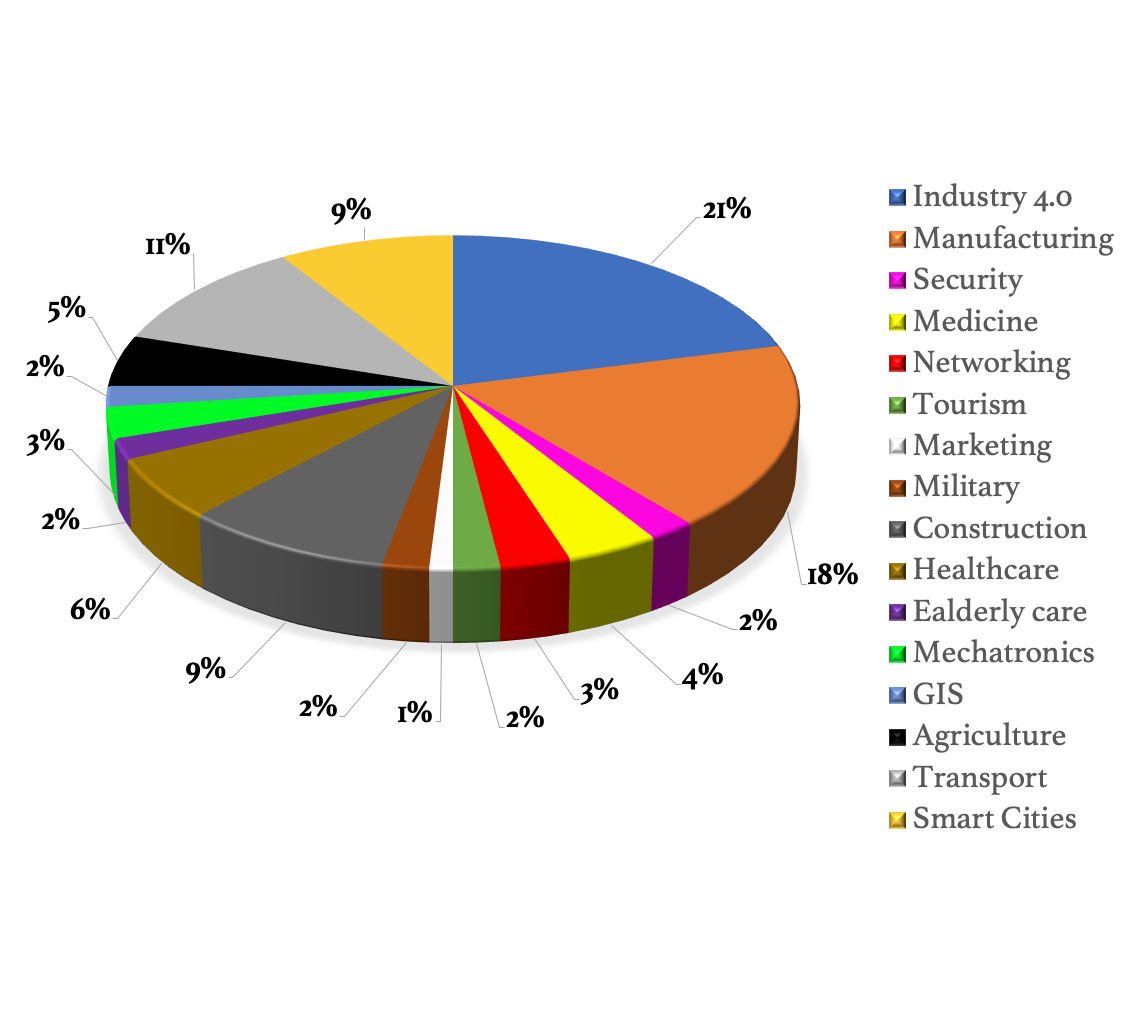}
  \caption{ Application areas of CPSS.}
  \label{fig:CPSS_application}
\end{figure}
Furthermore, in Figure \ref{fig:Co-authorship_network} we visualise the co-authorship network of the top 20 authors using VOSviewer software \cite{van2010software}. In addition to the above descriptive analysis, the co-authorship network provides further evidence that CPSS is not a niche topic of small community rather an emerging area of interest among may researchers from more than 30 countries across 5 continents. 

Since the goal of this study was also to identify the application areas of CPSS we did not restrict the search strings to specific domains. Consequently, the analysis revealed that despite adopting the CPSS notion, the selected papers address different issues in various application domains. In Figure \ref{fig:CPSS_application} we summarised a list of application areas where the concept of CPSS has been adopted according to the selected papers for these study.

\section{State of the Art Perspectives}
\label{sec:results}
In this section we present the state-of-the-art perspectives on CPSS by first exploring alternative terminologies and their corresponding definitions followed by a discussion on the CPSS paradigm with an exclusive analysis on works that used the CPSS acronym. Subsequently we present our analysis on how the social dimension/human aspect has been conceptualised in the CPSS literature.   

\subsection{Alternative Terminologies}
\label{sec:Alternative_terms}
\begin{figure}[!h]
  \centering
  \captionsetup{justification=centering}
  \includegraphics[width=0.7\textwidth]{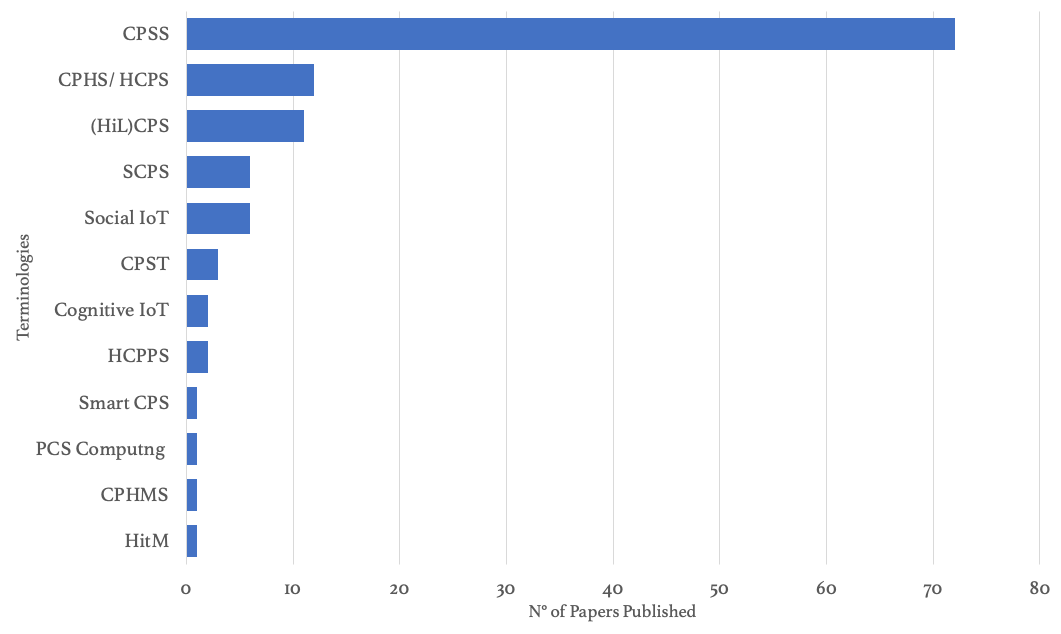}
  \caption{Distribution of terminologies over sample papers}
  \label{fig:Paper_terms}
\end{figure}

While studying the selected papers from Section \ref{sec:research_context:sota} we observed that a number of alternative terminologies to CPSS has been used by different researchers. Thus, we identified eleven terminologies for which a seemingly coherent definitions could be extracted. A summary of the  terminologies and their extracted definitions is  presented in Table  \ref{tab:definitions_Terminologies}. Figure \ref{fig:Paper_terms} illustrates the distribution of identified terminologies over the sample references used in this study.

\begin{table}[]
\renewcommand{\arraystretch}{1.3}
\caption{Alternative terminologies to CPSS and their corresponding definitions as proposed by different researchers .}
\begin{tabularx}{\textwidth}{p{.08\textwidth}p{.75\textwidth}X}
\toprule
 Term & Definition & Reference \\ \midrule
CPHS/ HCPS & Cyber-Physical-Human System (CPHS) or Human-Cyber-Physical System(HCPS) is a system of interconnected systems (i.e. computers, devices and people) that interact in real-time working together to achieve the goals of the system –which ultimately are the humans’ goals.  & \cite{liu2020human,Sowe2016, Smirnov2017a,kumar2017securing,  6685535,7961580,Zhang2017, Sharpe201937, 7844922, 8815669, Fantini2018, zhou2018toward, ZHOU2019624}   \\ \midrule
(HiL)CPS & Human-in-the-Loop Cyber-Physical System ((HiL)CPS) is a system consisting of a loop that involves humans, an embedded system (cyber component), and the physical environment where the embedded system augments a human interaction with the physical world making humans'  intents, psychological states, emotions, and actions an intrinsic part of any computational system. Thus, establishing a feedback control loop. 
& \cite{munir2013cyber, Schirner2013,6926397,7857283, Koenig2016, WANG2016377, Stankovic2016, JIRGL2018225,Nunes:2018:PIH:3217559,BAVARESCO2019109365, CIMINI2020258}\\ \midrule
SIoT & Social Internet of Things
(SIoT) is a kind of Social network where every node is an object capable of establishing social relationships with other things in an autonomous way according to rules set by the owner. SIoT is created by integrating social networking (SN) principles into the native IoT model. & \cite{7333282, 8016215, 8280086, 8425303, 8726793, Smart2019,7434330} \\\midrule
SCPS & Social Cyber-Physical System (SCPS) is  Cyber-physical system (CPS) that strongly interacts with the human domain and the embedding environment, working according to the expectations of humans, communities and society, under the constraints and conditions imposed by the embedding environment.
& \cite{horvath2012beyond, Yao2016, Kannisto2016, 8119375, 8000176, inproceedings}\\\midrule
CPST & Cyber-Physical-Social Thinking (CPST) is a concept emerged through the fusion of CPS and IoT on the basis of cloud computing technology, as a broader vision of the IoT. Precisely CPST is a hyperspace established by merging a new dimension of thinking space with the CPS.
 & \cite{Zhu2016a, NING2016504, 7428352}\\ \midrule
HCPPS & Human Cyber-Physical Production System (HCPPS) is a generic architecture with the control loop, adaptive automation control systems, and human-machine interaction to support humans, machines, and software to interface in the virtual and physical worlds so as to create a human-centric production system.
 & \cite{ZOLOTOVA2020105471, ROMERO2020106128}\\\midrule
 CIoT & Cognitive Internet of Things (CIoT) is a paradigm aimed at improving performance and to achieve intelligence of IoT through cooperative mechanisms with Cognitive Computing technologies that try to mimic human-like Cognitive capabilities, such as Understanding, reasoning and Learning. & \cite{dutta2018beyond,zhang2012cognitive} \\ \midrule
 HitM & Human in the Mesh (HitM) refers to human activities in Cyber-Physical production system in which the worker is participating to the process of production planning and its loop of control, and it is usually enacted by the role of the Manager.  & \cite{FANTINI2019133} \\\midrule 
 CPHMS & Cyber-Physical Human–Machine Systems (CPHMS) is a CPS  that includes problems of cognition (planning and decision making), navigation, action, human-robot interaction (perception, environment sensing, and interfacing with the end-user), and architecture development and middleware.
  & \cite{quintas2016information}\\\midrule 
  PCSC & Physical-Cyber-Social Computing is a paradigm that  encompasses a holistic treatment of data, information, and knowledge from the Physical, Cyber and Social worlds to integrate, correlate, interpret, and provide contextually relevant abstractions to humans.
 &\cite{Sheth2010} \\\midrule 
 Smart-CPS & is a CPS that combines various data sources (both from physical objects and virtual components), and applying intelligence techniques to efficiently manage real-world processes. & \cite{DELICATO20201134}\\
\bottomrule
\end{tabularx}
\label{tab:definitions_Terminologies}
\end{table}

Although there is a fundamental variation in nomenclature and design among these systems, many commonalities can be observed. For instance in terms of their components, all contain Human, Computers and Smart devices as component systems. Furthermore, they all share similar global objective (\textit{i.e.} despite resorting different techniques all systems aim to ensure a better human-machine synergy across various application areas). This is commonly shared by the other works that use the \textbf{CPSS} acronym.

  In the next subsection we present the results from exploration of the rest of the works in our sample references that adopt the CPSS acronym. particularly we present the core definitions given for CPSS, we then discuss the \textbf{CPSS paradigm} and take a closer look at how the social dimension (human aspect) is conceptualised in literature.

\subsection{Cyber-Physical-Social System (CPSS) }
\label{sec:CPSS}
 Our analysis revealed that the ways of defining CPSS is also different from one research work to another. Particularly, we identified five major categories of definitions that can summarise the understanding of the CPSS notion in literature presented in  Table \ref{tab:definitions_sample}. The definitions are categorised under the following themes inspired by the emphasis given by majority of the works: \textit{Command and Control, Social Sensing, Self-organisation, Big Data and Networking}).
\begin{table}[!h]
\renewcommand{\arraystretch}{1.3}
\caption{Definition categories of the CPSS acronym in literature.}
\begin{tabularx}{\textwidth}{p{.09\textwidth}p{.6\textwidth}X}
\toprule
Theme &  Definition & Reference \\ \midrule
Command \& Control & CPSS is a system consisting of a computer system, a controlled object and interacting social components (e.g., humans). Thus allowing the control of physical object using computation and social data to ultimately achieve moral goals and online cloud management of social processes. & \cite{Ganti2008senseworld, 6611338, Gharib2017, 8267947, murakami2012cpss,  8267950, Zhong2020, 8409112, Naudet2018, Gao2020, Xin2020, Liu2020, Song2020, doi:10.1080/17517575.2018.1470259, ISI:000447944400006, ISI:000414931500019,Liu2011}    \\\midrule
Social Sensing & CPSS is a system consisting of not only cyberspace and physical space, but also human knowledge, mental capabilities, and socio-cultural elements. Information from cyberspace interacts with physical and mental spaces in the real world, as well as the artificial space mapping different facets of the real world. 
 & \cite{ISI:000416244800001,  7530087, Yilma2019, ISI:000434003300001, ISI:000447348300008, ISI:000451598900059, ISI:000441187400008, wang2017transportation}
\\ \midrule
Self-organisation& CPSS is a system comprising three intertwining subsystems  \textit{(i) The human-based system} which refers to the social system containing human actors and their interconnected devices/agents and/or social platforms providing human-based services,
\textit{(ii) The software-based system} that refers to the cyber world providing software-based services including the underlying infrastructures and platforms, either on-premise or in the Cloud and  \textit{(iii) The thing-based systems} referring to the physical world that includes sensors, actuators, gateways and the underlying infrastructures. CPSS tightly integrate
physical, cyber, and social worlds to provide proactive and personalized services for humans.
 & \cite{Smirnov2014a, Guo2015327, smirnov2015cyber, Zeng2016, 7476853, 7557393, 7865933, 7862854, candra2016reliable, Dai2020, Huang2020, Wang2018, zhang2018cyber, Yilma, Wang2019a, Gan2020, Xu2020, liu2018towards, utomo2017usability}
\\\midrule

Big Data & CPSS is an extension of CPS/IoT formed by introducing human's social behaviour fostering a synergetic interaction between computing and human experience. Thus, integrating Big Data Collectors (BDCs), Service Organizers (SOs) and users to build a unified data-centric computing framework. 
 & \cite{Xiong2015320, Costanzo2016, CASSANDRAS2016156, 7444777, wang2010emergence, Bu2017, dautov2018data, zhang2018cyber, li2019multi, 8794542, Wang201942, Shen2020, Zhu2020, Peng2020a, Peng2020, Wang2020a, Wang2020, Jiao2020, Zhang2020, ISI:000418968400035, ISI:000455711200016, 8357923}
\\\midrule
Networking & CPSS is a  paradigm originates from the technology development of the cyber-physical systems (CPS) and cyber-social systems (CSS) to enable smart interaction between cyber, physical and social spaces, where {\bf CPS} includes communicators, multimedia entertainment and business processing devices, etc. and {\bf CSS} refers to social networks such as Facebook, Twitter, Youtube, etc.
 & \cite{Sheth2010, candra2016monitoring, mendhurwar2019integration, ISI:000452634800021, sisyanto2017hydroponic}\\
\bottomrule
\end{tabularx}
\label{tab:definitions_sample}
\end{table}

Despite the alternative terminologies used and their different conceptualisations discussed in Section \ref{sec:Alternative_terms} a common understanding of CPSS shared among all works can be summarised by the following definition. 
\begin{def1}\label{def:CPSSgeneric}
A \textbf{CPSS} is an environment cohabited by humans and smart devices that are in a virtual and physical interaction.
\end{def1}
\noindent
 Inherently the emergence of CPSS is tightly coined with the presence of human at the vicinity of CPS devices. We subsequently, we take a closer look at how human aspects, i.e., the the social dimension, has been conceptualised in CPSS literature.

\subsubsection{Conceptualising the Social dimension /Human aspects in CPSS}
\label{sec:human_aspects}
\noindent  Exploring the state-of-the-art we discovered that there are two main school of thoughts regarding the role of human in CPSS. Thus research works systematically adopt one of the two views discussed below:

\begin{enumerate}
\item 
\textit{Human as a sensor}: This is relatively the earliest view in the evolution from CPS to CPSS which originates with the increasing use of sensor-enabled smart devices by humans.  In this view the social aspect was brought by considering humans as sources of information for Cyber-Physical systems (\textit{i.e.} sensors). This conceptualisation primarily focuses on fusing various information originating from the social space (humans and their observations) with cyber-systems and physical-systems in order to accommodate various application needs.


\item \textit{Human as a system component}: 
On the other hand most of the recent works tend to conceptualize the social aspect of CPSS not only by considering human as a social sensor but also as co-creators being an integral part of the system. It is also known as the human-centric way of conceptualizing CPSS~\cite{Sheth2010}. This way of conceptualization considers humans as members of the CPSS, involving observations, experiences, background knowledge, society, culture and perceptions (i.e human intelligence and social organizations (e.g. Communities)) in order to co-create products and services together with the CPS. Here humans play the role of resources in that they provide information, knowledge, services, etc., which at the same time they consume, thus becoming users of the CPSS. 
\end{enumerate}    
 Adopting the second view in recent years, a considerable advance has been made in CPSS research. Particularly research works had successful results in designing smart environments and objects/machines to perform complicated tasks. These results are becoming more and more evident in various fields. According to~\cite{BOUFFARON201412317}, putting humans and machines to work closely  by promoting collaboration, learning and supervision can potentially deliver better outcomes than isolated operations. The pursuit of smartness in machines has allowed achieving high quality in task execution, in some cases even surpassing a human potential. However, achieving high quality in task execution by machines does not guarantee a seamless human-machine interaction. 

A human is a social creature and social interaction is by far the most seamless experience one can have as far as interaction goes. This is because it captures not only task related engagements of a human but also behavioural, emotional and cognitive characteristics which are deemed as the three layers of human response in any kind of interaction. \cite{Norman02}.  Furthermore, in the context of CPSS, humans experience mental as well as physical loads while interacting with CPS devices which is often a subjective experience for which people respond differently depending on individual skills, interests, preferences, etc. Since the CPSS paradigm considers humans as one components of the system, understanding social dynamics plays a crucial role to guide the integration of humans in CPSS.
Considering a human-to-human interaction we can consciously communicate our emotional, cognitive and behavioural responses because we are equipped with similar sets of sensors and information processing units \cite{Yilma2019,yilma2020new}. However, unlike a human counterpart detecting these aspects is not yet with in the reach of CPS or the so-called smart machines.

As we can understand from the state-of-the-art analysis, ongoing efforts in the CPSS paradigm are ultimately aimed at ensuring a human-machine synergy. Although this is the rationale behind many of the works, without a common definition of the concept and underlining principles to guide the integration of social aspects, it is evident that the CPSS paradigm is still in its infancy.  Therefore establishing a  common ground and some underlining principles of the concept are needed in order to facilitate such multidisciplinary collaborations and support the evolution towards a true CPSS. Following this, in the next section we propose a systemic formalisation of CPSS illustrated through a meta-model to establish a common ground based on the theory of systems and System-of-Systems(SoS) principles. 

\section{A Systemic formalisation of CPSS}
\label{sec:CPSS_formalisation}

CPSS is a complex system as it is composed of many components which may interact with each other; more precisely it is a System-of-Systems (SoS) \cite{Maier1996}. From the multiple theories on systems, our work takes as groundings the seminal work of Von Bertalanffy, one of the founders of the General Systems Theory (GST) \cite{1972_Von} and the System-of-Systems (SoS) principles from \cite{Maier1996,Morel2007SystemOE}. From these theories, we extract the general concepts and their relationships, which can be exploited as an ontological basis to describe CPSS as a system. In the following, we detail the main principles we use as a core systemic ground for defining and modelling CPSS.

\subsection{Definition of CPSS}
\label{sec:research_context:systems_theory}
 In the GST, a system is defined very generically as \textit{a complex set of interacting elements, with properties richer than the sum of its parts}~\cite{1972_Von}. In a more recent work on systems interoperability, defining an ontology also grounded in GST, but also on the work of others like Le Moigne~\cite{arnaud1991jean}
 , Naudet et al. \cite{NAUDET2010176} proposed a definition of system fitting more our context, which we reused for defining and modelling CPSS:
\begin{def1}\label{def:system}
"A \textbf{System} is a bounded set of interconnected elements forming a whole that functions for a specific finality (objective) in an environment, from which it is dissociable and with which it exchanges through interfaces" ~\cite{NAUDET2010176}.
\end{def1}

\noindent In this work the authors characterised a system by its components and the interactions between them, where each component can itself be a system. In this latter case, we can talk about a System-of-Systems (SoS), which is a concept that came into common usage in the late 90's to characterise large systems often formed from a variety of component systems which  developmentally and operationally exhibit the behaviour of complex adaptive systems (CAS) \cite{owens1996emerging, nla.cat-vn1543374}. The SoS notion fundamentally captured the non-monolithic nature of complex modern systems. The earliest and most accepted definition of SoS is given by Maier et al.~\cite{Maier1996} and is formalised as follows:

\begin{def1}\label{def:SoS}
 "A \textbf{System-of-Systems (SoS)} is an assemblage of components which individually may be regarded as systems, and which possesses two additional properties: operational and managerial independence of the components"~\cite{Maier1996}.
\end{def1}
\noindent Operational independence means that if the system-of-systems is disassembled into its component systems the component systems must be able to usefully operate independently, while managerial independence means the component systems not only can operate independently, but do operate independently: the component systems are separately acquired and integrated but maintain a continuing operational existence independent of the system-of-systems~\cite{Maier1996}. 
From a systems engineering perspective, the notion of SoS was best described as an emergent system from at least 2 \textit{loosely coupled systems}  that are collaborating \cite{Morel2007SystemOE}.
\noindent 
The SoS principle dictates that the relationship between component systems is recursive as any system is produced by another higher system, answering specific requirements. For a dedicated project, the target system (CPSS in our case) is the final produced system, in this recursively loop.

As explained in section \ref{sec:results}, majority of the works perceive CPSS as an environment formed through the interaction of humans and various kinds of CPS devices. Here each of the interacting entities are independent systems with operational and managerial independence. Thus, the newly formed system as a result of their interaction is a System of Systems(SoS) (\textit{definition \ref{def:SoS}}). The typical examples of CPSS are often the so-called \textit{Smart spaces} \cite{ROMERO2020103224} such as \textit{smart enterprises, smart buildings, smart homes, smart cities, etc.} Although current literature considers such kind of SoS as a CPSS, Social aspects are barely realised in the smart devices. Thus, the CPSS paradigm should set clear expectations of a true CPSS in terms of incorporating social aspects.  Following the notion of systems and SoS principles presented above, we propose a systemic definition of CPSS to create a domain independent formal understanding of CPSS. Thus, making it suitable for adaptability across different domains.This helps to create a shared vision in order to guide ongoing efforts and inspire the development of novel ones.  To this end we propose the following formal definition of CPSS.
 
 \begin{def1}\label{def:CPSS}
\textbf{Cyber-Physical-Social System (CPSS)}: is a system comprising cyber, physical and social components, which exists or emerges through the interactions between those components. A CPSS comprises at least one physical component responsible for sensing and actuation, one cyber component for computations and one social component for actuating social functions. 
\end{def1}

This formal definition stipulates the fundamental requirements for the emergence of a CPSS, which are the basic components (\textit{Cyber, Physical and Social}) and a relationship between these components. Furthermore, it characterises the expectation for a next generation of smart devices which is having a \textit{social component} that allows them to detect reason and objectify social interaction responses of a human (\textit{i.e.} emotional, cognitive and behavioral). Therefore, the smart devices themselves are expected to evolve and become a CPSS as a single system. Where as the complex SoS formed when these devices interact with humans is what signifies a true CPSS where social aspects are realised. This means in the evolution of the CPSS paradigm we can distinguish between two kinds of CPSS where one is a single independent system and the other is a SoS. The first is formed as a result of the interaction between the three fundamental components(Cyber, Physical and Social) of a single system while the later is formed when there is a social interaction  among  independent systems. 

Having established this ground regarding the expectations for a true CPSS, subsequently we discuss key systemic properties one needs to consider while designing CPSS as a single independent system as well as a SoS. 


\subsection{Systemic model for CPSS}
\label{sec-sys_model}
\begin{figure}[!h]
  \centering
  \includegraphics[width=0.5\columnwidth]{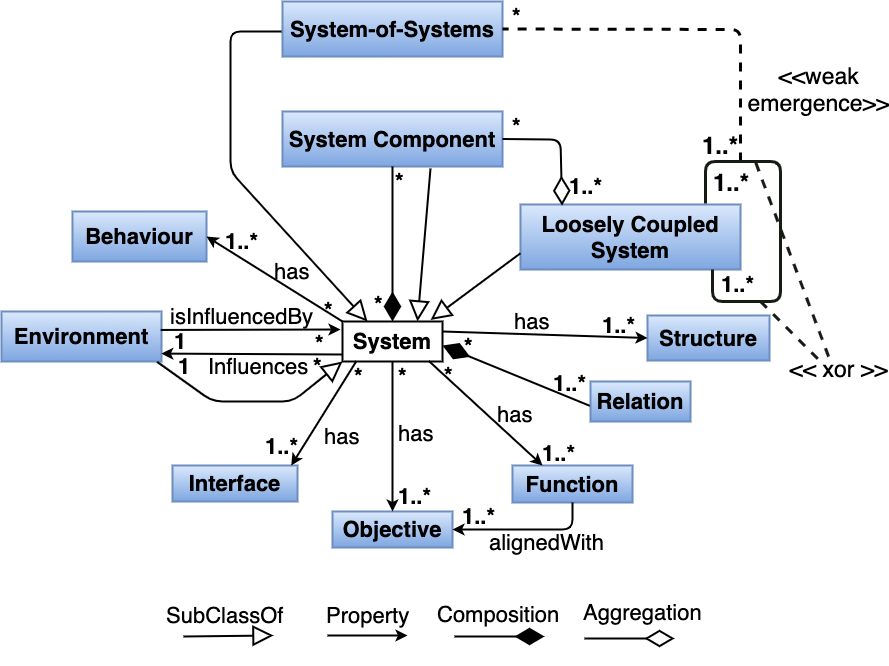}
  \caption{The Systemic model.}
  \label{fig:systemicModel}
\end{figure}

The complexity of any SoS  mainly depends on the nature of relations between its component systems, their individual behaviour, objectives and functionalities~\cite{Maier1996}. In a SoS each interacting entity being a system possesses all the following key systemic properties \textit{(Relation, Behaviour, Function, Structure, Objective, interface, environment and system component)}. Thus, framing CPSS as a SoS and aligning it with the theory of systems helps to ease the design process by allowing to clearly visualise the component systems, identify their individual objectives, relationships, inter-dependencies and  determine complementary as well as conflicting objectives. Subsequently on figure ~\ref{fig:systemicModel} we present a systemic model to be used as a basis to design a CPSS which depicts these key systemic concepts linking System and System-of-Systems (SoS). The systemic model is adopted from \cite{NAUDET2010176} and modified incorporating the SoS notion from systems engineering perspective by using UML 2.0 notation. 

The emergence of CPSS is inherently the result of \textit{relations} among components. Thus, using the systemic notion as a basis in the following we present a meta model of CPSS illustrating the emergence of CPSS and the types of relations forming a CPSS and other kinds of SoSs in a CPSS context.


\subsection{Meta-Model of CPSS}
\label{sec:meta-Model}
\begin{figure}[!h]
  \centering
  \captionsetup{justification=centering}
  \includegraphics[width=0.55\textwidth]{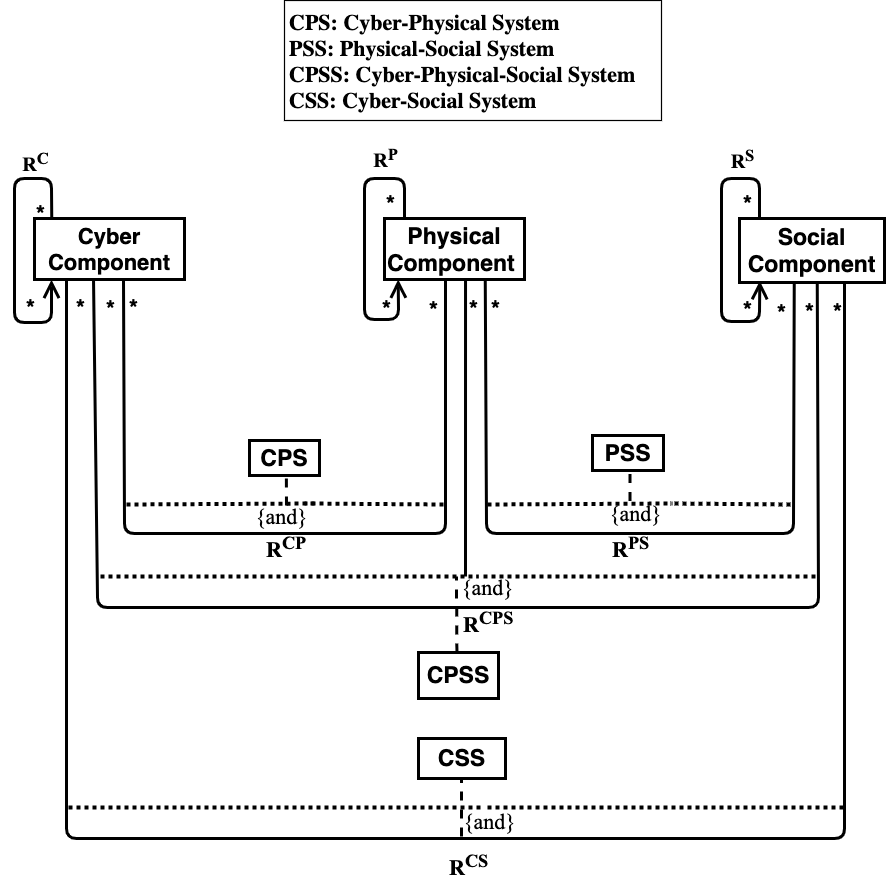}
  \caption{ The CPSS Meta-model.}
  \label{fig:meta-Model}
\end{figure}

 An earlier version of CPSS meta-model was proposed by \textit{Yilma et al.}\cite{Yilma2019} as an extension from \textit{Lezoche et al.}  \cite{lezoche2018cyber} to introduce the social component in CPS. In this subsection we propose two meta-models to illustrate the emergence of CPSS as system and SoS by using UML 2.0 notation.  The first meta-model presented in Figure \ref{fig:meta-Model} is an extension and a generalisation of these preceding works. It formalises the main components of a CPSS as combinations of fundamental (C)yber, (P)hysical and (S)ocial elements, as well as the relation between them. It allows representing the different kinds of systems that emerge when relations are instantiated: Cyber-Physical-Social System (CPSS), Cyber-Physical System (CPS), Physical-Social System (PSS), and Cyber-Social System (CSS).   

Let $C$, $P$, $S$ be respectively the set of Cyber, Physical, and Social components, and $R$ be the set of existing relations, in a system of interest. We define three kinds of relations:
\begin{itemize}
 \item $R^X$: \textit{X $\times$ X $\rightarrow$ R}, where \textit{X} is ``C", ``P" or ``S"
 \item $R^{XY}$: \textit{X $\times$ Y $\rightarrow$ R}, where \textit{X} and \textit{Y} are``C", ``P" or ``S" and X $\neq$ Y
\item $R^{CPS}$: \textit{C $\times$ P $\times$ S $\rightarrow$ R}
\end{itemize}
There are in total seven types of relations that can link components of a CPSS together, which we detail in the following: $R^{C}$, $R^{P}$, $R^{S}$, $R^{CP}$, $R^{PS}$, $R^{CS}$, and  $R^{CPS}$. 

\begin{itemize}
    \item $R^C$:- refers to a connection between cyber components, existing in the virtual space, for example an information flow, a command, query, etc.  It can also refer to the sharing of a computational resources. \textit{e.g,} two software packages sharing the same processing unit. 
    \item $R^P$:- refers to a connection between physical components, existing in the physical space. An example could be the connection between mechanical parts of a machine.
    \item $R^S$:- refers to a social relationship between social components. It can materialise as an information flow or a transfer of knowledge between social components. It also reflects cognitive ties that govern human behaviour, \textit{e.g.}, an intellectual conversation between people.  
    \item $R^{CP}$:- refers to a relationship that exists between cyber components and physical components that can potentially result in the integration of computation with physical processes (sensing or actuation), \textit{e.g.}, the relation between components of a smartphone to function. 
    The $R^{CP}$ relation   
    leads to the emergence of a CPS (Cyber-Physical System):
    \begin{equation}
        \forall C, \forall P, \exists R^{CP} \Leftrightarrow \exists CPS 
    \end{equation}
    
    \item $R^{PS}$:- refers to a relationship that exists between physical components and social components that can potentially result in cognitive processes and observable social behaviours. This is the property that enables a human to take actions that reflects his emotion, cognition and behaviour in a given context. 
    The $R^{PS}$ relation leads to the emergence of a PSS (Physical-Social System):
    \begin{equation}
        \forall P, \forall S, \exists R^{PS} \Leftrightarrow \exists PSS 
    \end{equation}
    
    \item $R^{CS}$:- refers to the relationship between Cyber and Social components that can potentially result in the integration of computation and social capabilities, \textit{e.g.}, virtual representation of people in a social network. 
    The $R^{CS}$ relation leads to the emergence of a CSS (Cyber-Social System):
    \begin{equation}
        \forall C, \forall S, \exists R^{CS} \Leftrightarrow \exists CSS 
    \end{equation}
  
    \item $R^{CPS}$:- refers to a relationship that exists between at least one cyber, one physical and one social component, that can potentially result in the integration of sensing, actuation, computation and social processes. 
    The $R^{CPS}$ relation is what glues the three components together leading to the emergence of a CPSS as an independent system.

    Formally, we define the emergence of a CPSS as a single system by the following formula:
    \begin{equation}
        \forall C, \forall P, \forall S, \exists R^{CPS} \Rightarrow \exists CPSS \label{eq:CPSS} 
    \end{equation}
\end{itemize}


The CPSS meta-model is built on top of the systemic meta-model presented in Section~\ref{sec-sys_model}. Formally, all component classes (\textit{Cyber Component}, \textit{Physical Component}, \textit{Social Component}) are subclasses of \textit{System Component}, and all system classes (\textit{CPS}, \textit{PSS}, \textit{CPSS}, \textit{CSS}) are subclasses of the general System class.  As systems, the latter inherit from all the properties detailed in Section~\ref{sec-sys_model}.
While the concepts of CPS and CPSS were already known, the meta-model introduces two relatively new concepts: PSS and CSS. \textit{PSS} is a composition of physical and social elements, where the social part is materialised through the physical part. The main representative is the \textit{Human} system: the physical part is the physical body, while the social part is composed of the attributes that generate social responses such as cognition, behaviour and emotion, that are observed through physiological changes on the body \cite{PERUZZINI2018105600}. The reader could argue that for human, the social system is indeed a part of the physical system. However, we view the intangibles of social system separately to study and better understand social aspects which we eventually want to transpose to machines. Although human is our topic of interest, the PSS is a super class encompassing also other living things with a social behaviors. \textit{CSS} corresponds to a system where the social component is manifested through the cyber component. A typical kind of CSS is a \textit{Social Network},~\cite{doostmohammadian2019cyber}, where the social activities actually result from interactions in the virtual world.
For a better readability in Fig.~\ref{fig:meta-Model}, relations are represented by a link, but all should be understood as subclasses of the systemic \textit{Relation} class from Fig. \ref{fig:systemicModel}. The constraint \textit{\{and\}} is used to represent the mandatory requirement of at least one component from each part in relation in order for a new system to emerge.

\begin{figure}[!h]
  \centering
  \captionsetup{justification=centering}
  \includegraphics[width=0.55\textwidth]{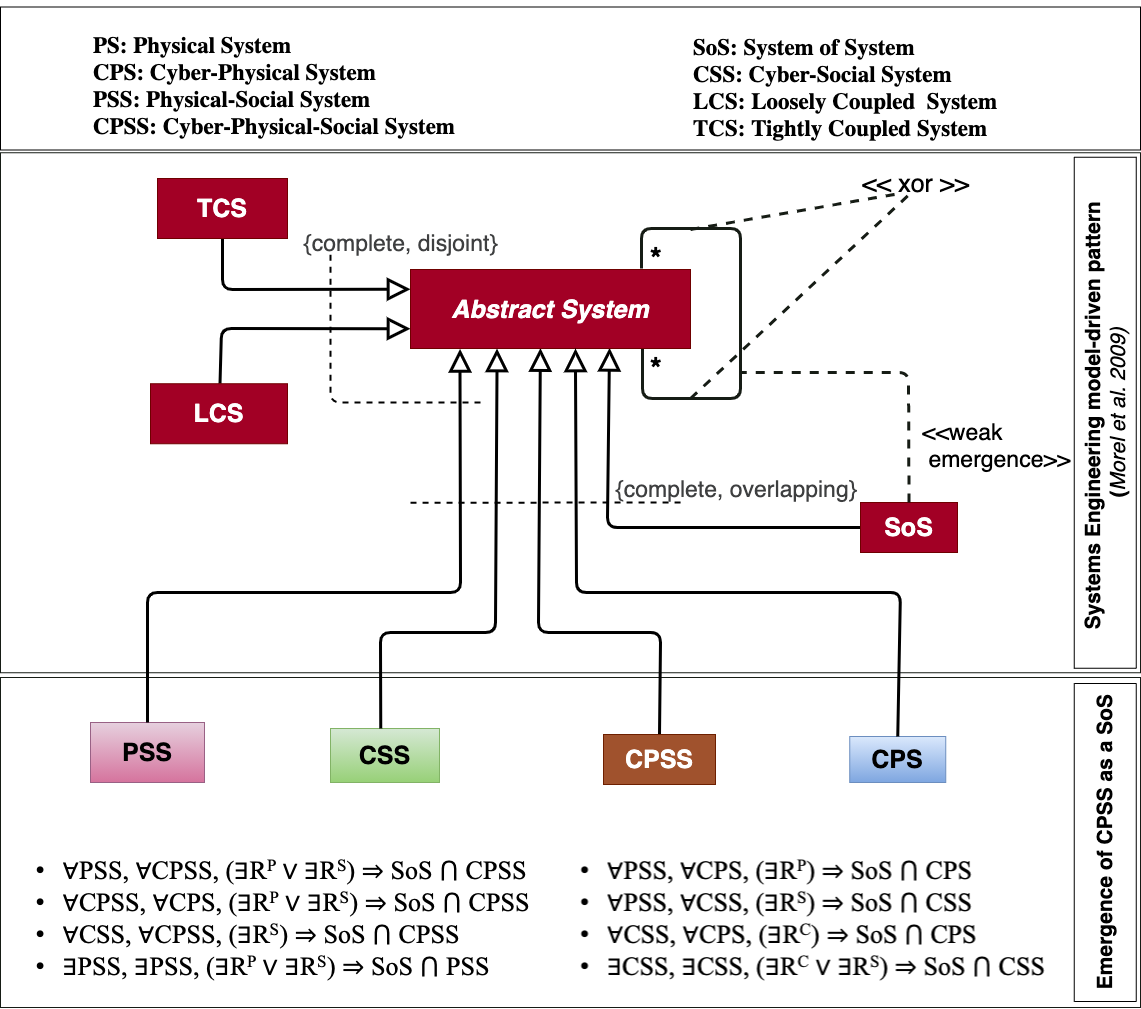}
  \caption{Extended Meta-model of CPSS as a SoS.}
  \label{fig:CPSS_SOSmeta-Model}
\end{figure}

\begin{figure}[!h]
  \centering
     \captionsetup{justification=centering}
  \includegraphics[width=0.45\textwidth]{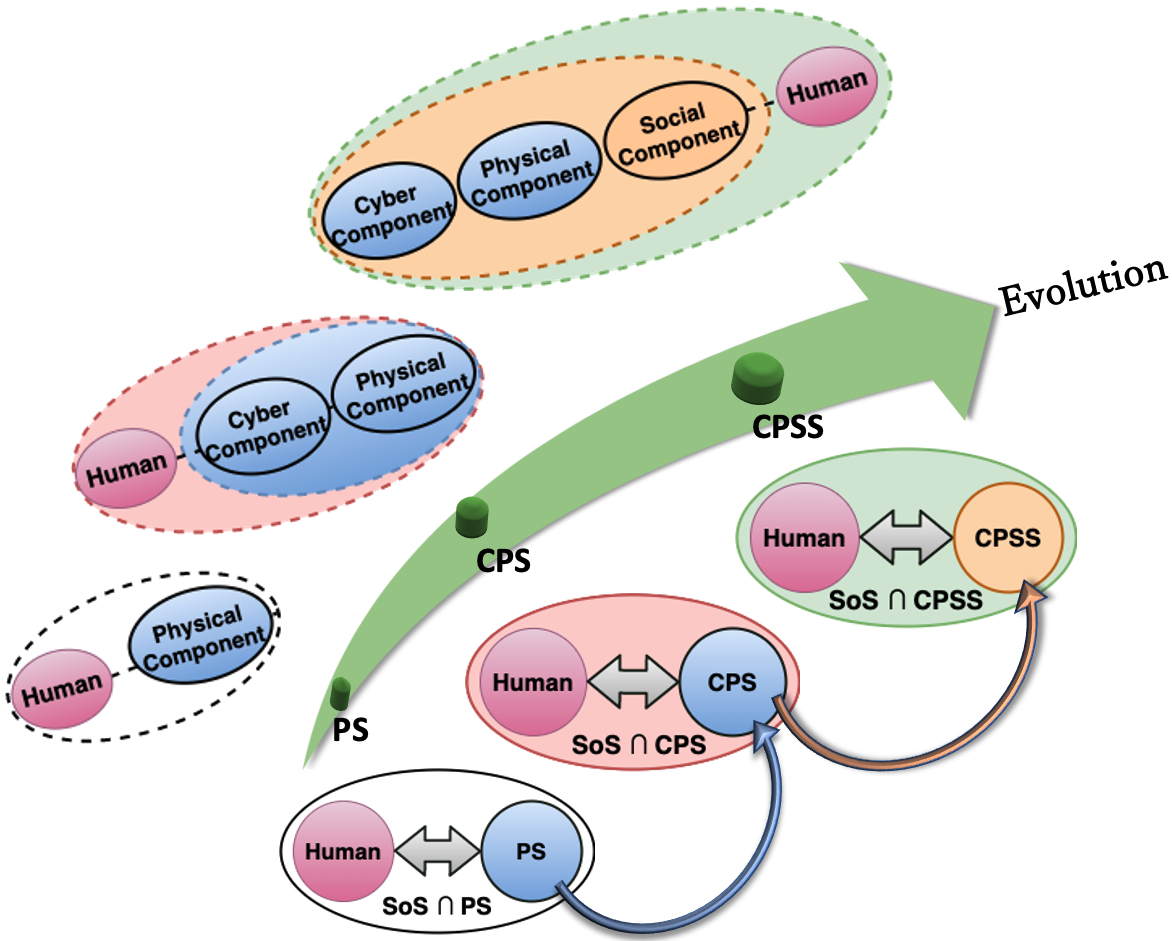}
  \caption{ CPSS Evolution.}
  \label{fig:CPSS_evolution}
\end{figure}

In order to visualise the emergence of CPSS as a SoS and also other types of SoSs formed as a result of the interactions between component systems, we present an extended meta-model in figure \ref{fig:CPSS_SOSmeta-Model}.
  As it can be seen on the meta-model the top part illustrates concepts adopted from the work of \textit{Morel et al.} \cite{Morel2007SystemOE} showing the formation of SoS as a weak emergence from the interactions between independent systems, that can be either  Tightly Coupled System(TCS) or Loosely Coupled System(LCS). The interaction link on abstract system refers to any of the relations discussed above. The bottom part shows the emergence of CPSS as a SoS and also other kinds of SoSs formed in CPSS context. The axioms at the bottom illustrate the main kinds of SoSs that can be formed as a result of interactions between the independent systems. Fundamentally the postulate here is that a true CPSS is formed as a SoS when there is a social relation $R^{S}$ between a single system CPSS \textit{e.g. Cobot(Collaborative robot)} and a PSS \textit{e.g. human}. Here, having a physical relation  $R^{P}$ instead of social $R^{S}$ can form a SoS. However, it does not necessarily entail the formed SoS is a CPSS which essentially requires a social relation $R^{S}$ where the single CPSS \textit{e.g. Cobot} is able to detect, reason and adapt to social interaction responses of the human. Furthermore, CPSS can also emerge as a SoS whenever a CPS or a CSS initiate  a social relation with a single system CPSS. The first 3 axioms on Fig. \ref{fig:CPSS_SOSmeta-Model} represent the basic ways a CPSS can be formed as a SoS. The rest of the axioms describe other kinds of SoSs that can be formed in a CPSS context.  
 Reflecting back on the state-of-the-art the evolution of CPSS is summarised in Figure \ref{fig:CPSS_evolution}. The bubbles below the arrow represent the composition of the SoSs formed as the components mature from \textit{Physical system(PS)} to \textit{CPSS}. Whereas the bubbles above the arrow depict the details in the make up of the individual systems that a human interacts with. As it can be seen in Fig. \ref{fig:CPSS_evolution} the current understanding of CPSS corresponds to the middle one, where we have a human interacting with socially constrained \textit{CPS devices}. However, our postulate is to eventually arrive at a true \textit{CPSS} where a human interacts mainly (but eventually not only) with the CPSS's social component which is materialised through the socially capable \textit{CPSS devices} represented by the top bubble. Hence, the \textit{CPSS} paradigm we propose in this work primarily aims at shading light in this direction to achieve social capability of machines.
 
The proposed formalisation of the CPSS paradigm  shows that the current state of the research is way behind from achieving the required level of maturity to be called a true CPSS. Hence, it opens opportunities in the CPSS research domain to propose solutions that can pave the way towards a true CPSS. From the state-of-the-art analysis it is also understandable that there are a number of open challenges to be addressed in the CPSS research such as \textit{resource management, interoperability, security, etc. \cite{Yilma}}. Nevertheless,  in the road towards a true CPSS, human dynamics remains being one of the major challenges  yet to be tackled.  Following this, in the next section  we present some perspectives and future direction we believe are feasible to pursue based on the proposed formalisation. 
\section{Perspectives and Future Direction}
\label{sec:CPSS_Future}
In this era of digitisation, where virtual workplaces are becoming a common trend, the popular opinion, and fear is that machines will continue to take ever larger portions of human work activities eventually replacing us. Although, the progressive changes especially in industry 4.0 seem to imply that we are on track towards full automation, there are still a wide range of opportunities to reimagine digital workplaces in the context of human-machine collaboration. As opposed to a race against one another we can redesign these systems blending human-machine participation to perform far more efficiently than either could individually. The ultimate vision of the CPSS paradigm shares this notion of fostering a seamless human-machine collaboration by instrumenting the human and socialising the machine \cite{Carrozza2019}. Nowadays as more and more people are becoming users of wearables and sensory devices, the leap in the concept of "quantified self" opens opportunities to instrument humans by taking advantage of the humongous amount of collected data. Although humans are still being instrumented for various purposes in the realm of CPSS, extrapolating true social dynamics for the socialisation of machines is yet to be explored. In the subsection below we discuss our perspectives on the implication of socialisation in the context of CPSS. Subsequently we discuss personalisation in CPSS as an example of a feasible future direction to contribute for the socialisation of machines.

\subsection{Towards Socialisation in CPSS}
\label{sec:socialisation}
Socialisation is a lifelong process through which we develop our personalities and human potential and learn about our society and culture through interactions\cite{clausen1968socialization}. We are surrounded by people who become part of how we act and what we value  which dictate norms and cultures in social interactions. In the digital world we live in today we are not only being surrounded by people but also by machines. As we cohabit our environments and also interact with the machines they are also part of our social circle, hence they possess the potential to influence our personality. Nevertheless, unlike a human counterpart machines are not yet able to detect and reciprocate complex emotional, cognitive and behavioral aspects. Therefore, as we continue to surround ourselves with more and more smart devices, empowering them to socialise so that they can respect important human values and needs  becomes a necessity.  This helps to avert the trajectory towards full automation and foster the creation of collaborative digitised environments where machines become better companions to co-create with and serve humans.  

In a nutshell the CPSS paradigm which introduces the \textit{Social} part  aims at creating an enhanced version of smart spaces where the current socially constrained \textit{"CPS devices"} gradually evolve to socially capable \textit{"CPSS devices"}. The process of socialising machines brings humans into the equation to learn from and interact with, since the true socialisation process in humans itself is a lifelong process of learning through interaction. In CPSS environments the availability of various kinds of sensors allows us to instrument humans for learning by analysing sensory data. Even though this is within the reach of current technology, extrapolating meaningful and unquantifiable social aspects from quantified data is a rather challenging task. We believe that this is where the grand challenge in the quest towards the socialisation of machines lies.  Particularly in the context of CPSS socialisation can take up on two important implications. The first one has to do with the learning process which exploits social dynamics to detect and reason social interaction responses of a human which are difficult to directly detect and analyse for machines. The second has to do with interactions which empower machines with social components to respond in a desirable manner (\textit{i.e. "Social actuation"}). We believe that addressing these challenges calls for a collaborative research effort among fields that investigate human dynamics such as \textit{Cognitive science, behavioural science,  affect/emotion recognition and related sub-fields of Artificial intelligence (AI)}. Especially recent advances made with the help of AI to indirectly infer emotional, cognitive and behavioral aspects of humans from physiological data analysis are promising to exploit social aspects  \cite{dinh2020stretchable}. These are often specialised fields dealing with specific aspects of human dynamics such as emotion, cognition, behavior, etc. Bringing these fields together the CPSS paradigm opens possibilities to design machines that closely mimic human-like social interactions in each of these aspects and continually learn to evolve with changing environments and human needs. 

Another important aspect to be noted in socialisation is that the uniqueness of personalities. Although social norms are often commonly shared among members of a social group, personal experiences and knowledge further shape individuals' preferences, interests and perception of their environment. Hence, peoples' actions and behaviours  are the reflections of their unique personalities. This is a determining factor for the quality of their interaction experience. Therefore, socialising machines should be able to recognise personal preferences, interests, limitations and opportunities of individuals in order to ensure a seamless interaction experience. This positions  \textit{Personalisation} at the heart of the CPSS paradigm. Personalisation by itself is a well evolved research field, and we believe the introduction of  \textit{Personalisation in CPSS} gives birth to an interesting topic  that can serve as a test bed for interdisciplinary efforts to tackle challenges of socialisation in CPSS.  As this is unexplored direction among others, particularly from the CPSS perspective discussed in this paper, in the next subsection  we layout some basic steps that we believe will guide the development of CPSS through personalisation as one feasible future direction. 
\subsection{Personalisation in CPSS}
 The field of Personalisation is a relatively older research field dating back to the late 90s. The notion of personalisation, which is broadly known as customization, refers to tailoring a service or a product in a way that it fits to specific individuals' preferences, cognition, needs or capabilities under a given context \cite{karlgren1990algebra}.  Virtual assistants on devices such as Alexa, Siri and Cortana,  Chat-bots, online recommendations for e-commerce and entertainment are among the popular areas where virtual personalisation has gained momentum.

Unlike  the virtual settings, a CPSS faces more complexities from the physical world. This is because there are a number of co-existing stakeholders. Each stakeholder has its own objectives to be satisfied, constraining each other  while at the same time being constrained by environmental factors. Furthermore, the actions and behaviours of people in CPSS are often the result of their individual preference, interests and other natural or environmental factors which are not yet fully explored. We believe introducing personalisation in a CPSS addresses these unique aspects of individuals and opens new possibilities to better manage the environments and improve quality of interaction experiences. Thus, personalising and making such a complex environment adaptable to humans essentially requires to  appropriately position the task of personalisation by taking into account not only the user of personalisation but also every entity in the CPSS environment that has a direct or indirect influence on the user as well as the personalisation). As we recall from the discussion in section \ref{sec:CPSS_formalisation}  every stakeholder(i.e component) in CPSS context is an independent system. 
Therefore, the problem of personalisation in CPSS can be formalised as a function of the main component systems (\textit{i.e.} the user $u$ of personalisation service, the CPSS in which she evolves in $cpss$, the crowd of other people in the CPSS $cr$, the smart device that implements the personalisation service $d$ and the global context $cx$) can be written as: 
\begin{equation}\label{eqn:PuCPSS}
Perso_u^{(CPSS)} = f(u,cpss,cr,d,cx)
\end{equation}
Here, the context $cx$  refers to the set of all other elements (component systems) of the CPSS \textit{e.g.}$\{x_{1}, x_{2}, ... x_{n} \}$ that have no direct or indirect influence on the user/personalisation. When any of the component systems in $cx$ has an impact on the user/personalisation it will be taken as part of the formulation $f$ as $f(u,cpss,cr,d,x_{i}, cx)$ ; $\forall$ $x_{i}$ $\in$ $cx$.

Let us consider a scenario of Cobotic system in a \textit{smart workshop} setting. In smart manufacturing systems, we have  engineers, operators and maintenance technicians that are skilled and able to perform tasks on different machines. In these settings Cobots are introduced at workshops to collaborate with the workers in order to improve efficiency. However, Cobots are often programmed only to execute predefined tasks. Hence, they are not able to adapt to changing needs of human workers. This can potentially degrades collaboration quality and could also compromise safety of human workers. Therefore, the primary benefit to introduce personalisation here would be enabling cobots to understand and reason dynamic human interaction responses and adapt to changing personal needs accordingly.
In doing so, the personaliser should also take into account their co-existing objectives and respect systemic rules. Thus, adopting the global formulation of personalisation in CPSS (equation \ref{eqn:PuCPSS}), the problem of personalisation in Cobotics can be formulated as a function of the main systems (\textit{i.e.} the user of personalisation service translates to the worker  $w$, the CPSS which translates to the smart workshop $ws$, the crowd of other people in the factory translates to a team of workers $tw$, the device implementing the personalisation which translates to the Cobot $cb$ and the context elements $cx$) can be written as :
\begin{equation}\label{eqn:PuCBx}
Perso_u^{(Cob)} = f(w,ws,tw,cb,cx)
\end{equation}

\begin{figure*}[!h]
  \centering
  \includegraphics[width= 0.8\textwidth]{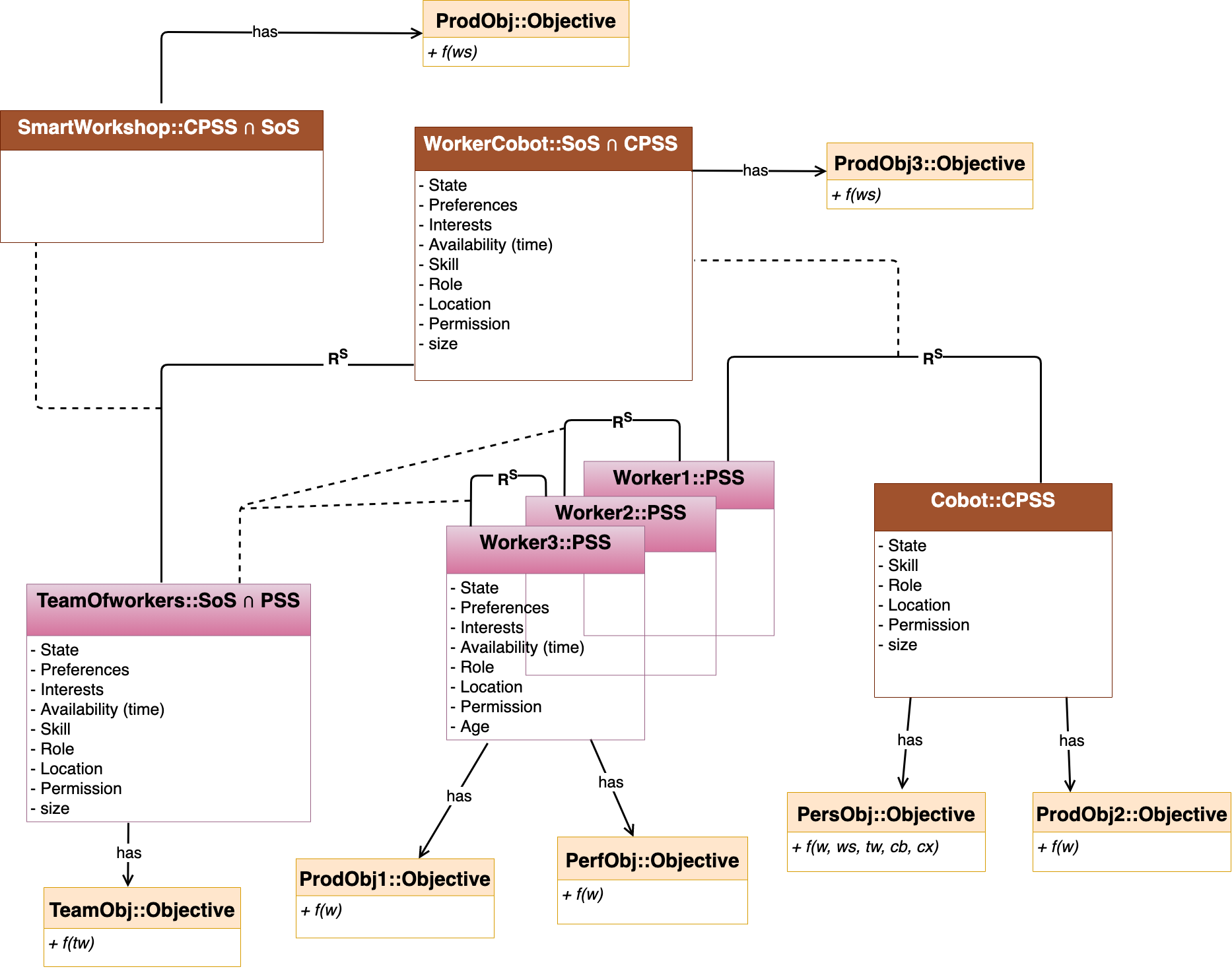}
  \caption{Conceptual model of a Smart factory based on the CPSS meta-model}\label{fig:fw}
\end{figure*}

Following this formulation in figure \ref{fig:fw} we present an  instantiation of the meta-models presented in section \ref{sec:meta-Model} towards the specific scenario of smart workshop.
 As it is depicted on  figure \ref{fig:fw} the class \textit{Cobot} is instantiated as a subtype of \textit{CPSS} provided a personalisation objective (\textit{PersObj})  and a production objective(\textit{ProdObj2}). The class \textit{Worker} is an instance of \textit{PSS}. Whereas the class \textit{WorkerCobot} represents a CPSS which is a SoS that emerges as a result of the relations between a worker(PSS) and a Cobot(CPSS) according to axiom 1 on figure \ref{fig:meta-Model}.  
 The class \textit{TeamOfworkers} is another emergent SoS formed as a result of relations among two or more workers. The class   \textit{SmartWorkshop} is thus, a CPSS which is a SoS formed from \textit{TeamOfworkers} and \textit{WorkerCobot} relations.  As an independent system each of these systems can have one or more objectives serving the global purpose of the smart workshop as well as personal ones. This aids an initial design process by giving better visibility to key systemic properties of every entity.  Consequently the personaliser needs to make the best possible compromise/ trade-off between the objectives. This may as well translate to a constrained multi-objective optimization problem.
 
In this particular scenario of a smart workshop personalisation is one objective which is implemented by the Cobot interacting with a worker. This essentially means enabling the Cobot to learn complex human interaction responses. Thus, gradually adapt to changing states respecting important human values and needs to become better companions. Implementing this however is not a trivial task as it requires relaxing the control rules and training cobots to derive efficient representations of the humans state from high-dimensional sensory inputs, and use these to generalize past experience to new situations. 
Such kinds of challenging tasks are remarkably solved by humans and other animals through a harmonious combination of reinforcement learning(RL) and hierarchical sensory processing systems \cite{serre2005object, fukushima1982neocognitron}. This in particular has inspired the development of several RL algorithms over the years \cite{nguyen2020deep} used for training agents to perform complicated tasks. However, their application  was limited to domains in which useful features can be handcrafted, or to domains with fully observed, low-dimensional state spaces. Recently a novel artificial agent called deep Q-network (DQN) was proposed in the work of \cite{mnih2017methods}. DQN can learn successful policies directly from high-dimensional sensory inputs using end-to-end reinforcement learning. DQN has been tested over various complicated tasks and was able to surpass the performance of all
previous algorithms \cite{silver2016mastering, silver2017mastering}. It has also enabled the creation of ``AlphaGO";which is to date considered as one of the greatest breakthroughs in artificial intelligence that was able to beat the world's  most diligent and deeply intelligent human brains \cite{chen2016evolution}. This and other recent successes such as ``ÀlphaStar" \cite{10.1145/3319619.3321894}  demonstrate the potential of RL to build intelligent agents by giving them the freedom to learn by exploring their environment and  make decisions to take actions which maximises a long term reward.We believe that RL can be beneficial to the task of personalisation in CPSS as it allows agents to learn by exploring their environment unlike supervised methods which require collecting huge amount of labeled data and harder to train with continuous action space. Thus, in our context RL can be used to find optimal policies on which the Cobot operates on in order to take the best possible action given the state of the worker. While this seems an efficient strategy to relax control rules allowing cobots to interact by exploration, the grand challenge of inferring the intangible social interaction responses remains yet to be solved. 
Although such responses are often hard to directly detect and analyse, as mentioned above successful results in the domains of \textit{Affect recognition, Cognitive science and Behavioral science} can be jointly leveraged to make a step in the socialisation of machines.

In general formulating the personalisation task in such a way based on the systmic notion and the meta-model allows us to clearly visualise and profile every entity, their individual objectives, relationships and interdependence. This strategically positions CPSS to benefit from innovative and multidisciplinary solutions in tackling social dynamics. Therefore, we are optimistic that relying on the premises established in this work, a multidisciplinary approach is a worthwhile endeavour in the quest towards a true CPSS.

\section{Conclusion}
\label{sec:conclusion}

This work presented a Systematic Literature Review on Cyber-Physical-Social System (CPSS) exploring the state-of-the-art perspectives regarding definitions, application areas and conceptualisations of the social aspect. The analysis revealed that the way of defining CPSS among researchers has been inconsistent. Furthermore, there are no common principles to guide the integration of social aspects in CPSS. Following this a systemic formalisation of CPSS was proposed  based on the theory of systems and System-of-Systems principles. Particularly a formal definition of CPSS was proposed to characterise fundamentals of CPSS and expectations for next generations of smart systems in terms of integrating social aspects. This was done to establish a common ground and guide the development of novel methods to incorporate the full spectrum of social aspects into CPSS. The definition was then supported by two meta-modes illustrating the emergence of CPSS in two different forms (\textit{i.e. }as a device and as an interaction space). Having established a vision for the future of CPSS, the formalisation reveals that the current conceptualisations are not where they need to be to arrive at the required level of maturity. Going forward the formalisation and the meta-model are believed to open opportunities for multidisciplinary efforts to gradually introduce social aspects in CPSS research.  To this regard personalisation was proposed as one feasible future direction within the reach of current technologies. In general the foundation established in this work aims to inspire and support  innovative directions to attenuate issues of social dynamics in CPSS to ultimately ensure human-machine synergy.  

\bibliographystyle{elsarticle-num}
\bibliography{cas-refs}

\end{document}